\pdfoutput=1

\documentclass[preprint]{acmart}
\usepackage{forest}
\usetikzlibrary{angles}
\usepackage{multicol}
\usepackage{subcaption}
\usepackage{tcolorbox}

\def\BibTeX{{\rm B\kern-.05em{\sc i\kern-.025em b}\kern-.08emT\kern-.1667em\lower.7ex\hbox{E}\kern-.125emX}}

\copyrightyear{}
\acmYear{}
\setcopyright{none}
\acmConference{}
\acmBooktitle{}
\acmPrice{}
\acmDOI{}
\acmISBN{}

\begin{document}

\title[Automated Search for Configurations of DNN Architectures]{Automated Search for Configurations of Deep Neural Network Architectures}

\author{Salah Ghamizi, Maxime Cordy, Mike Papadakis, Yves Le Traon}
\affiliation{
  \institution{SnT, University of Luxembourg}
}
\email{email: {firstname.lastname}@uni.lu}





\renewcommand{\shortauthors}{Ghamizi, et al.}

\begin{abstract}
Deep Neural Networks (DNNs) are intensively used to solve a wide variety of complex problems. Although powerful, such systems require manual configuration and tuning. To this end, we view DNNs as configurable systems and propose an end-to-end framework that allows the configuration, evaluation and automated search for DNN architectures. Therefore, our contribution is threefold. First, we model the variability of DNN architectures with a Feature Model (FM) that generalizes over existing architectures. Each valid configuration of the FM corresponds to a valid DNN model that can be built and trained. Second, we implement, on top of Tensorflow, an automated procedure to deploy, train and evaluate the performance of a configured model. Third, we propose a method to search for configurations and demonstrate that it leads to good DNN models. We evaluate our method by applying it on image classification tasks (MNIST, CIFAR-10) and show that, with limited amount of computation and training, 
 our method can identify high-performing architectures (with high accuracy). We also demonstrate that we outperform existing state-of-the-art architectures handcrafted by ML researchers. Our FM and framework have been released 
 to support replication and future research.

\end{abstract}




\maketitle

\section{Introduction}
\label{sec:introduction}

Nowadays, Deep Learning (DL) systems are used in a wide variety of domains as decision-making tools. 
Given an input (e.g., radiograph image, front view of a car, etc.), they are able to \emph{classify} this input in a specific category or \emph{class} (e.g. presence or absence of cancer tumour, appropriate steering angle, etc.). 
Such solutions offer self-learning capabilities: when \emph{trained} with large amounts of labelled data (i.e., inputs with their associated output class), they automatically learn classes and predict, with some confidence, the classes of unseen inputs. This makes DL systems easy to use (require limited human intervention), flexible (adapt to many use cases), and effective (achieve high accuracy). 


A DL system often consists of a classification model displaying the form of a \emph{Deep Neural Network} (DNN), which is a network of neurons that includes the \emph{input layer} (which encodes the characteristics of the given input), the \emph{output layer} (which represents the classes to which the input can be classified) and, in-between, multiple intermediate \emph{hidden layers}. For example, to classify inputs with 10 characteristics into 4 classes, we can use a 3-layer DNN, which is composed of one hidden layer that is connected both to the 10 neuron input layer (one input neuron per characteristic of the input) and to 4 neuron output layer. 

From the above example, one can easily see that DNNs have parameters to configure. Usually, the number of neurons in the input and output layers are determined by the problem instance, but the number and shape of hidden layers are choices to be made by engineers. By adding more layers, one can create DNNs capable of solving complex tasks with potentially higher accuracy, yet at an increased computation costs (when training the model) and with a higher risk of \emph{overfitting} (i.e., being accurate on training data but failing to classify well unseen data). In practice, there exist multiple alternative layer shapes that can be used, which adds complexity to the configuration process. 

To develop appropriate DNNs, engineers resort on intuition, experience, trial and error, or generic models (models already developed for another purpose). This process is usually tedious, non-systematic and potentially sub-optimal (the resulting architectures may not achieve best performance). To deal with this issue, we propose an automated search-based method that identifies prominent DNN architectures for given problems. More precisely, we rely on variability modelling techniques to represent the configuration space of DNN architectures, and configuration search techniques \cite{Similarity2014, Henard2015} to search for optimal architectures within this space.

We start by modelling the variability within DNN architectures in a Feature\footnote{The term ``feature'' refers to a variation point in a variability system, and should not be confused with the the input space characteristics that are usually referred to in ML.} Model (FM)~\cite{Kang1990} extended with attributes and feature cardinalities~\cite{Czarnecki2005b,Benavides2010,Michel2011,Cordy2013}, where each valid configuration (aka variant, product) of our FM corresponds to a deployable DNN architecture. 
We show that sophisticated FMs are convenient to represent the variability of DNNs by demonstrating how to derive, from a FM, DNN architectures that solve image recognition problems, e.g., LeNet5\cite{LeCun1998GradientbasedLA} and Inception modules.

We then implement a mapping from our FM to Tensorflow and Keras, which are standard programming libraries for Neural networks. Thus, from any valid configuration, we can automatically generate an implementation of the corresponding architecture in these libraries. 
On top of the above framework we apply search algorithms to generate configurations according to some strategies, e.g., maximizing configuration diversity. To do so, we use PLEDGE~\cite{Henard:2013:Pledge}, a configuration generation tool, and link it with Tensorflow+Keras. All in all, our framework automatically selects, builds and evaluates the accuracy of candidate architectures on specified datasets.

We evaluate our method on two image classification problems, which involve two commonly used datasets (MNISt and CIFAR-10). We use the LeNet5 and SqueezeNet (state-of-the art DNN architectures) as a baseline for comparison. Our results show the capability of our method to explore the configuration space and to search for good architectures. Notably, we manage to find architectures that outperform LeNet5 and SqueezeNet in raw accuracy on both datasets.

Overall, our work paves the way for applying software product-line and configuration techniques to DL systems. We offer a practical and feasible way to automatically optimize DNNs by specializing their configurations on particular domains. We, thus, deemed our endeavour successful and expect future research to contribute with novel configuration search techniques tailored to this context. To support this task, we make our models and framework publicly available.


\section{Motivation and Research Questions}
\label{sec:rqs}



Our goal is to provide engineers with a tool to configure and specialize (to specific tasks) DNNs. To achieve this, 
we rely on a variability model that captures all the variation points of DNN models together with the validity constraints. As typically performed in configurable systems, (valid) configurations selected from the variability model are linked to specific implementations (in our case, libraries such as Tensorflow and Keras) to derive deployable DNNs. Based on this framework, engineers can train and evaluate the resulting DNNs according to their needs. 

We aim at representing the whole variability of DNNs with a FM. In this first endeavour, we  
focus on the architecture of DNNs (i.e. how their neurons are structured and connected) and leave aside the training process (training rate, cost function, gradient descent method, etc.) and the dataset preparation (e.g. data augmentation) since the architecture space is the most complex part 
(see Section~\ref{sec:variability} for more details). 
In view of this, our first research question is:

\begin{tcolorbox}[colback=white]
    \textbf{RQ1:} Can we develop a variability model that represents all possible DNN architectures?
\end{tcolorbox}

Having formed the variability model, we turn our attention on the configuration process. We aim at specializing DNNs to particular classification tasks by exploring and evaluating the performance of selected configurations. As the configuration space is large and constrained, we rely on techniques that have been successfully used to generate configurations of configurable systems. Our second research question is thus formed as:


\begin{tcolorbox}[colback=white]
    \textbf{RQ2:} Can we effectively search the configuration space and identify well-performing DNN architectures?
\end{tcolorbox}

Ultimately, the benefits of our configuration search is the identification of high-performance architectures. This means that specializing a DNN architecture to a particular task should lead to optimal or nearly-optimal results.  We evaluate this by comparing the automatically generated architectures with state-of-the-art DNNs. Thus, we ask the following question:

\begin{tcolorbox}[colback=white]
    \textbf{RQ3:} Does our technique finds DNN architectures that outperform the state of the art?
\end{tcolorbox}

\section{A Variability Model for DNNs}
\label{sec:variability}

While DNNs are classically viewed as a linear sequence of fully-connected neuron layers (often named \emph{dense layers}), new layer structures have been introduced over the past twenty years. For example, Convolutionnal Neural Networks (CNNs)\cite{LeCun1998GradientbasedLA} -- mostly used in image recognition -- have introduced the convolution layer and the pooling layer. The former produces a set of patterns generated by convolving successive filters on an input signal~\cite{KrizhevskyCNN2012}, thereby emulating the behaviour of the visual cortex. To do so, it applies matrix operations to merge multiple sets of information. On the other hand, a pooling layer aggregates an input matrix into a smaller one to reduce computation cost and overfitting. 

\begin{figure}
\vspace{-1.0em}
    \centering
    \includegraphics[width=\linewidth]{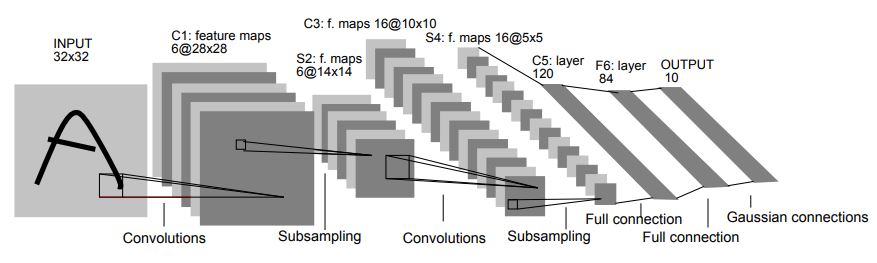}
    \caption{LeNet5 architecture as traditionally viewed}
    \label{fig:LeNet5}
    \vspace{-1.0em}
\end{figure}

One of the most established DNN architectures based on convolution layers is LeNet5, of which Figure~\ref{fig:LeNet5} depicts the structure as presented originally by LeCun et al. \cite{LeCun1998GradientbasedLA}. Its input is a matrix (usually of pixels) which is analyzed through a three-step process. Each of the first two steps involve a convolution layer and a pooling layer (named subsampling in Figure~\ref{fig:LeNet5}). The last step consists of going through one convolution layer and one dense (fully-connected) layer that, in the end, performs the classification itself.

More recently, Inception Network~\cite{Szegedy2015GoingDWI_Inpception} architectures popularised new constructs like branching and merging of layers, while Residual Network (ResNet)~\cite{He2016DeepRL} architectures can make neurons' output connections skip the next layers to propagate their computation farther in the network. All those advanced mechanisms make the classical, over-simplistic way of seeing DNNs inappropriate to represent the variability of their architectures. 

This raises the need for a generalized representation allowing the combination of fine-grained constituents of any of the above layer structures. We believe that FMs are an appropriate modelling formalism to this end. Indeed, their compositional form allows the definition of a hierarchical structure where concepts are further and further decomposed, while providing the ability to assemble a holistic system from those inner constituents. Moreover, since their introduction by Kang et al.~\cite{Kang1990}, they were extended over the years to model the variability of an increasing variety of systems. 

\begin{figure}
\vspace{-1.5em}
\centering
\includegraphics[width=9cm]{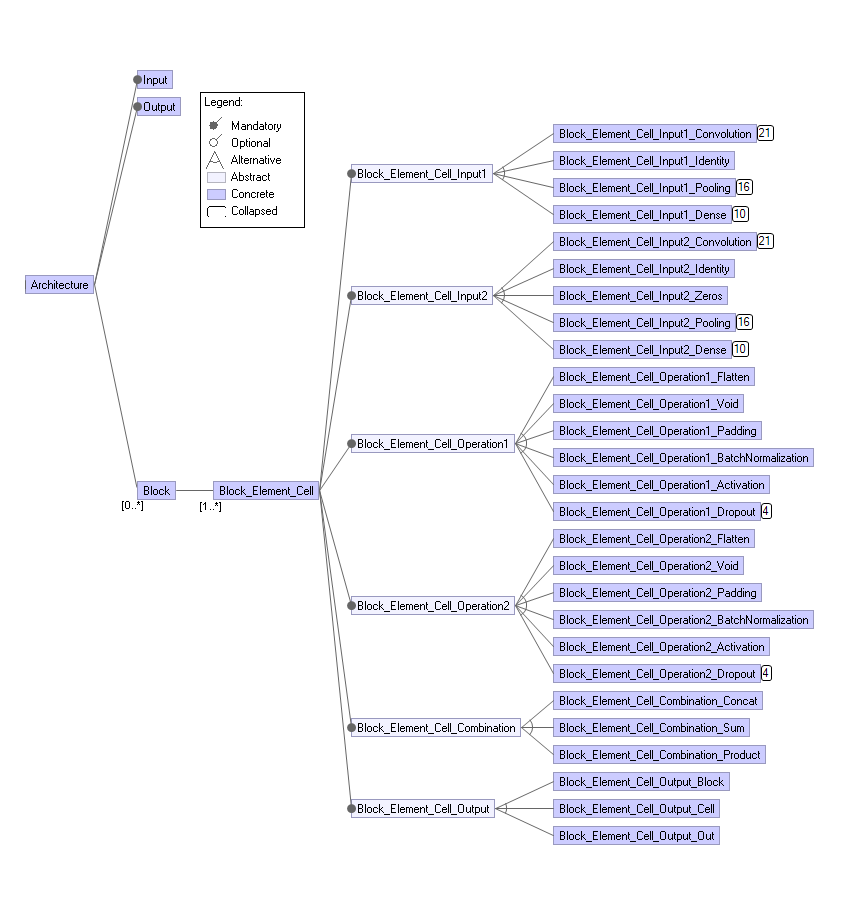}
\caption{FM of DNN architectures, with feature attributes collapsed. The FM has been built using FeatureIDE \cite{Thum2014}. }
\label{fig:FeatureNet-Graph}
\vspace{-0.5em}
\end{figure}

Figure \ref{fig:FeatureNet-Graph} shows our FM of DNN architectures designed using FeatureIDE~\cite{Thum2014b}. An architecture consists of two mandatory features representing the \texttt{Input} structure (corresponding to the initial matrix input to the DNN) and the \texttt{Output} structure (corresponding to a dense layer with a softmax activation function and whose number of neurons is the number of classes), as well as a multi-feature \texttt{Block}.\footnote{A multi-feature is a feature with a maximum cardinality greater than one.} The instances of \texttt{Block} together represent the hidden layers of the DNN, lying between the input layer and the output layer. Each instance includes one child multi-feature named \texttt{Cell}, which represents the particular structure we invented and that generalizes over the traditional structure of neuron layers. 

\begin{figure}
\centering
\includegraphics[width=6cm]{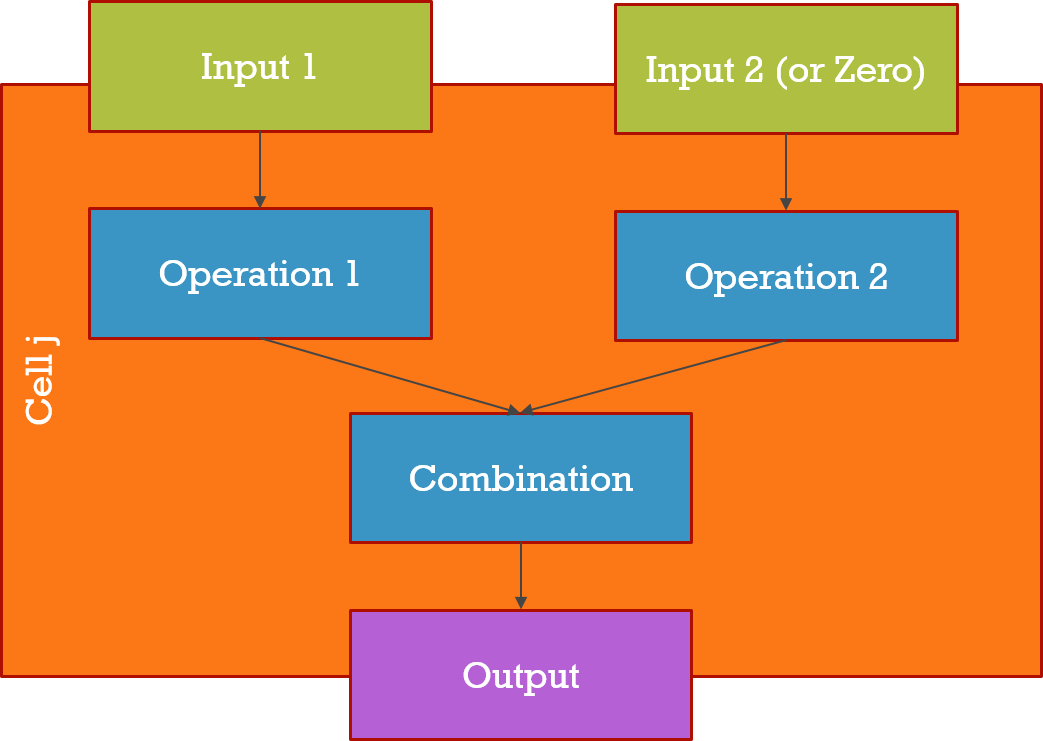}
\caption{The inner constituents of a cell.}
\label{fig:FeatureNet-Cell}
\vspace{-1.0em}
\end{figure}

\begin{figure*}
\centering
\includegraphics[width=\linewidth]{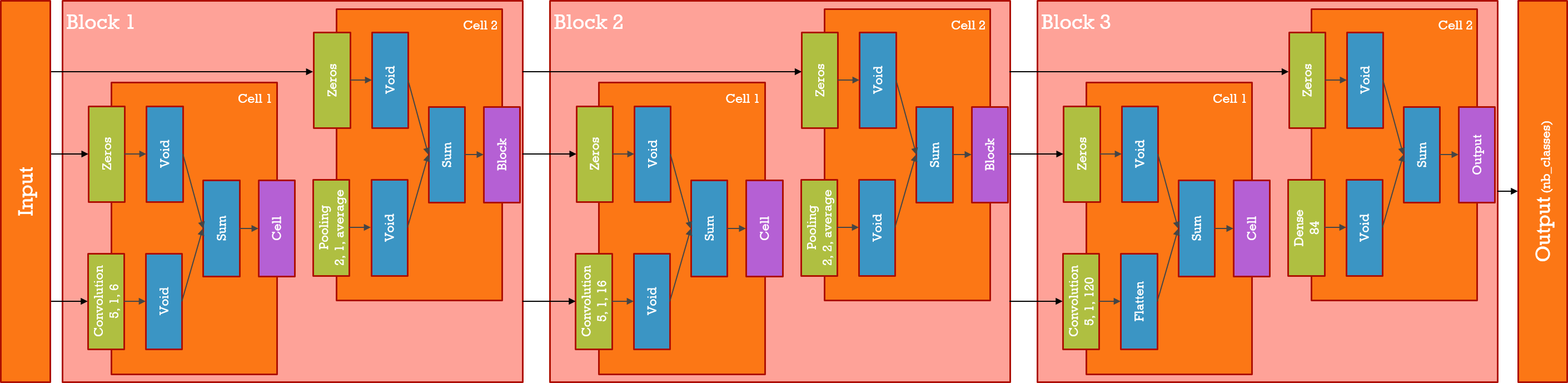}
\caption{A LeNet5 architecture modelled using our representation. It contains three blocks where the first two are Convolution-Pooling steps and the last block operates the reasoning with Dense Layers}
\label{fig:FeatureNet-LeNet5}
\end{figure*}

The structure of a cell is better illustrated on Figure~\ref{fig:FeatureNet-Cell}. It is constituted by five elements, each of which corresponds to a child feature of \texttt{Cell} in the FM. \texttt{Input 1} and \texttt{Input 2} both represent a layer (named an input layer of the cell) that receive an input (i.e. a matrix) to the cell and reason on it. \texttt{Input 2} is in some sense optional, as it can be filled with zeros to simulate that the cell accepts one input. \texttt{Operation 1} and \texttt{Operation 2} are operations applied to the results of the two input layers in order to prepare their combination. \texttt{Combination} defines what type of combination is performed. Finally, \texttt{Output} defines how far in the network the result of the result of the combination must be sent. 

Before detailing each of those constituents, we first give an example of an architecture modelled as blocks and cells. In Figure~\ref{fig:FeatureNet-LeNet5}, we illustrate the use of our cell-based representation to model the LeNet5 architecture previously introduced in Figure~\ref{fig:LeNet5}. Each of the three blocks corresponds to one of the three steps performed by a LeNet5 DNN. In each of the first two blocks, we find two cells: one corresponding to the convolution layer and the other one to the pooling layer. The third block has two cells: one for the convolution layer and the other for the dense layer. Since in LeNet5, the input simply goes from one layer to another, all cells use zeros as second input layer, \emph{void} operations (i.e. no operation), and combine results by summing them.


\textbf{Inputs}. In addition to Zeros layer, input layers can be of four types, represented by alternative features in Figure~\ref{fig:FeatureNet-Graph}. \texttt{Identity} simply keeps the input of the cell as it is. \texttt{Dense} corresponds to a layer of fully-connected neurons. It has two attributes: \texttt{neuron\_number} designates the number of neuron in the layer, while \texttt{activation} denotes their activation function (e.g. sigmoid, softmax). 

\texttt{Pooling} aggregates meaningful information from the cell's input. It operates on a matrix and can either extract the average or the maximum of every square areas of that matrix. The size of the areas is defined by its attribute \texttt{kernel}, while the \texttt{type} attribute defines whether to extract the \texttt{average} or the  \texttt{maximum}. Pooling layers can also skip elements of the original matrix and the amount of skipped elements after each extraction is defined by the \texttt{stride} attribute.
Last, the attribute \texttt{padding} defines how to deal with parts of the extracted area that are out of the original matrix, i.e. filling the missing parts with zeros or discarding the whole area.


\texttt{Convolution} corresponds to a convolution layer and has multiple attributes. \texttt{Kernel} is the size of the filters that will be used, where a filter is a matrix that allows the convolution to detect different patterns (primitive patterns for images include, e.g., edges or corners). Additionally, the \texttt{filters\_number} attribute denotes how many filters this matrix will try to learn. The \texttt{stride} and \texttt{padding} attributes are defined as in \texttt{Pooling} inputs, while \texttt{activation} denotes the activation function used by the layer. 

\textbf{Operations}. The most common operation is \texttt{Void}, meaning that it does not alter the result of the input layer. Another is \texttt{Flatten}, which converts the result matrix into a vector. \texttt{Padding} adds zeros to the matrix to change its size, while \texttt{Activation} denotes a specific activation function used to modify the value boundaries of matrix elements. Some operations can also be used to introduce regularization like \texttt{Dropout} -- which simulates the random dropping of neurons with a probability rate given by the feature attribute \texttt{rate} -- or \texttt{BatchNormalization}, which normalizes the activations of the input layer at each batch. 

\textbf{Combination}. There are three types of combination: \texttt{Sum} simply sums the results, \texttt{Concat} concatenates them, while \texttt{Product} perform a dot product matrix operation.

\textbf{Output}. We distinguish between three types of destination to which a cell's output is sent. It can be a subsequent cell of the same block (i.e. \texttt{Cell output}) or the next block (i.e. \texttt{Block output}), or the output is one of the final outputs of the network (i.e. \texttt{Out output}). A \texttt{Cell output} has an attribute named \texttt{relativeIndex}, which indicates that the output in transmitted to the $(relativeIndex + 1)$th next cell.

This description of the constituents of a cell completes our definition of the FM's features. In addition to the feature decomposition groups, cross-cutting constraints define validity rules for the architecture. Due to lack of space, we do not represent them as logic formulae but, instead, we explain them intuitively. First, the $i$-th block/cell can only exist if the $(i-1)$-th exists as well. Besides, if the output of a cell is a \texttt{Block output}, there must exist a next block. If it is a \texttt{Cell output}, there must be at least a number of subsequent cells equal to $relativeIndex + 1$. Relative indexes must also be defined such that a cell is not the target of more than two outputs. Additionally, Pooling and Convolution layers are only executed on matrices whose size is compatible with their kernel size. Flatten operations cannot be applied on 1-dimensional matrices. Combinations (sum, concat and product) must be performed on two matrices with compatible size.

\textbf{LIMITATIONS}. Overall, our FM encompasses a large class of feedforward neural networks. It does not, however, support Recurrent Neural Networks (RNNs), which would require looping connections between neurons. This restriction is due to the linear structure of our blocks and cells. Extending our modelling to support RNNs is feasible, but would require (i) allowing cells' output to reach a preceding cell or block and (ii) introduce new input layers, operations and combination.

\section{Implementation}
\label{sec:implementation}

\subsection{Forming Architectures}

Many automated techniques have been proposed to generate/select configurations of variability-intensive systems. Some of them attempt to cover all (1-wise, pair-wise, and more generally t-wise) feature interactions \cite{Johansen:2012:AGT:2362536.2362547}. In this work, we use diversity (in terms of enabled/disabled features of the FM)~\cite{Henard:2013:Pledge, Similarity2014} as a driving criterion. Intuitively, the generation method should allow the formulation of architectures with diverse shapes (in terms of layers). Such techniques have been intensively studied and compared in the area of configurable systems \cite{Medeiros:2016:SamplingComparison, Varshosaz:2018:Samplingstudy}. Here, we aim at demonstrating that it is possible to represent and search the DNN architecture space so that we can find useful configurations. Therefore, we choose diversity as it has been shown to be effective at global space exploration \cite{abs-1709-06017, Similarity2014}. 

More precisely, we use the PLEDGE tool~\cite{Henard:2013:Pledge}, which generates a given number of configurations by using a search algorithm that maximizes the diversity of the provided sample. Here, the term diversity refers to the mean distance (on the feature space) between any two configurations of a given sample. 
By feeding our FM into PLEDGE, we can thus sample a diverse set of DNN architectures.

The drawback of PLEDGE is that it does not handle feature cardinalities and feature attributes. Therefore, we rely on the array-based semantics of extended FMs proposed by Cordy et al.~\cite{Cordy2013} to (i) transform numeric and enumerated attributes into alternative Boolean features, (ii) unfold each multi-feature in a corresponding sequence of single features with the same underlying structure, (iii) transform all constraints accordingly. 
This process resulted in an increased size of our FM which, e.g., includes 3,296 features and 8,004 constraints when both the numbers of blocks and cells per block are limited to 5.

\subsection{Deploying DNN architectures}

We have developed a prototype tool (named \emph{FeatureNetCompiler}\footnote{publicly available at https://github.com/yamizi/FeatureNet}) on top of the Tensorflow and Keras frameworks. It takes as inputs configurations (selected by PLEDGE) and generates the corresponding DNN models in Tensorflow+Keras. Those models are obtained by building the DNN architectures defined by the configurations and using default training parameters. Our tool then trains and tests the generated models on a given dataset. 

A particular difficulty we faced when implementing our tool was the handling of cells' output that are not transmitted to the next cell. To achieve this, we designed a data structure to store the pending cell outputs. In this structure, each cell's output is stored together with a number equal to its \texttt{relativeIndex} attribute. Initially, this number represents the distance between the cell that has produced the output and the cell that will consume it. An output is said to be \emph{active} when its associated number reaches zero. Then, a newly reached cell uses the active elements, or the inputs of the block if there is no such elements. As active elements of the heap are used, they are removed from the data structure and the associated number of all remaining elements is reduced by one. This mechanism allows the merging of multiple chains into one, and supports complex branching, skipping and merging, which is required to model, e.g., an Inception Module~\cite{Szegedy2015GoingDWI_Inpception}.



\section{Evaluation protocol}
\label{sec:setup}

\subsection{Methodology}

RQ1 aims at evaluating the conformance of the FM we presented in Section \ref{sec:variability}. To answer this RQ, we demonstrate that we can form existing and arbitrary deployable architectures. In particular we check whether we can form architectures from 1,000 configurations produced by PLEDGE and two manually-configured architecture (LeNet5~\cite{LeCun1998GradientbasedLA} and Inception~\cite{Szegedy2015GoingDWI_Inpception}). We thus check whether we can deploy and train these architectures. To further validate our model, we also check the semantic equivalence between the LeNet5 architecture generated by our tool and the manually-built counterpart of LeNet5 (implemented directly in Tensorflow/Keras). This is checked by comparing the average accuracy over multiple runs (on both training and test sets) during the model evolution (i.e. the training epochs). 
Note that we do not check semantic equivalence on Inception, as it would require excessive computing power (multiple weeks to train it with the same experimental protocol). 


RQ2 concerns the capability of our approach to search for well-performing architectures. To evaluate this, we form architectures based on 1,000 configurations produced by PLEDGE. We train these architectures for 12 epochs, on the selected datasets, and compute their accuracy. Since the configurations are diverse, we expect to see a wide range of accuracy values. 

RQ3 compares our architectures with the state of the art. To enable a fair comparison, we use LeNet5 (as parameterized in \cite{LeCun1998GradientbasedLA}) and SqueezeNet \cite{Iandola2017SqueezeNetAA} as baselines. These two architectures have a size (expressed in number of weights) of the same order of magnitude as the median size of architectures we generate. 
We compare the accuracy and the efficiency of the all the architectures on the datasets, when trained for 12, 300 and 600 epochs. 
In case our generated architectures achieves better accuracy and efficiency than LeNet5 and SqueezeNet, this would mean that the cumbersome process of manually constructing efficient DNN architectures can be automated.

\subsection{Experimental Setup}


\textbf{Forming architectures}. Unsurprisingly, preliminary experiments revealed that training a large number of large-sized architectures requires tremendous computation resources. In order to keep the required computation time at a reasonable experimental level (less than 24 hours) we applied two restrictions. First, we limited the configurations generated by PLEDGE to 1,000. Second, we added the following constraints in our FM:

\begin{itemize}
  \item The maximum number of blocks and the maximum number of cells per block are set to 5.
  \item Convolution layers include at most 128 filters, each of which has a maximum size of 5x5.
  \item Dense layers have at most 512 neurons each.
  \item Batch normalization and padding operations are forbidden.
  \item Concatenation and product combinations are forbidden.
  \item There can be only one \texttt{Out output} (more would typically be used to improve training).
\end{itemize}

The above constraints allow keeping training time under a reasonable threshold, even though some generated architectures still take more than 2 hours, when trained for 600 epochs. However, it prevents us to generate more costly architectures like ResNet \cite{He2016DeepRL}, Inception, and AlexNet \cite{KrizhevskyCNN2012}. Still common and even some recent architectures remain available, in particular LeNet5 and SqueezeNet, which we use as baselines in our experiments.

\textbf{Training setup}. In all our experiments, we rely on a commonly used setup for training. We apply stochastic gradient descent with categorical crossentropy as loss function, a learning rate of 0.01 without any decay or momentum, and a batch size of 128. We did not try to optimize these hyperparameters, as they are likely architecture-dependent and we wanted to focus only on sampling architectures and evaluating them on a common ground.

\textbf{Datasets}. We consider two datasets of image recognition problems that are widely used in research on machine learning. The first dataset is \emph{MNIST}, which is composed of 28x28 greyscale images, each of which represents one of the ten digits. We used 60,000 images for training and 10,000 images for test. The second dataset concerns \emph{CIFAR-10}, a collection of 60,000 labelled images scattered in 10 classes, including 50,000 for training and 10,000 for testing.

\textbf{Performance metrics}. Our experiments make use of two performance metrics that are commonly used to evaluate DNN models. The first one is the classical accuracy, which is defined as the ratio of images that are well-classified by the DNN model. A higher accuracy thus means more correct predictions. The second is the \emph{efficiency} of the model, that is, the accuracy divided by the total number of weights computed in the model, which is an indication of the resources required to train it. Hence, efficiency represents a tradeoff between correctness of predictions and computation resources.

\textbf{Hardware}. We run PLEDGE on a laptop with CoreI7-8750H processor and 16GB RAM. The training and testing of neural networks was performed on a Tesla V100-SXM2-16GB GPU. 





\section{Results}
\label{sec:results}

\subsection{RQ1: conformance of Feature Model}

To check the conformance of our FM, we successfully deployed, trained and test (with the related datasets) 1,000 configurations generated by PLEDGE, as well as LeNet5 and Inception. This fact shows that our model leads to valid architectures.

\begin{figure*}
\vspace{-1.0em}
\centering
\includegraphics[width=0.9\linewidth, trim={1cm 1.2cm 0cm 0cm},clip]{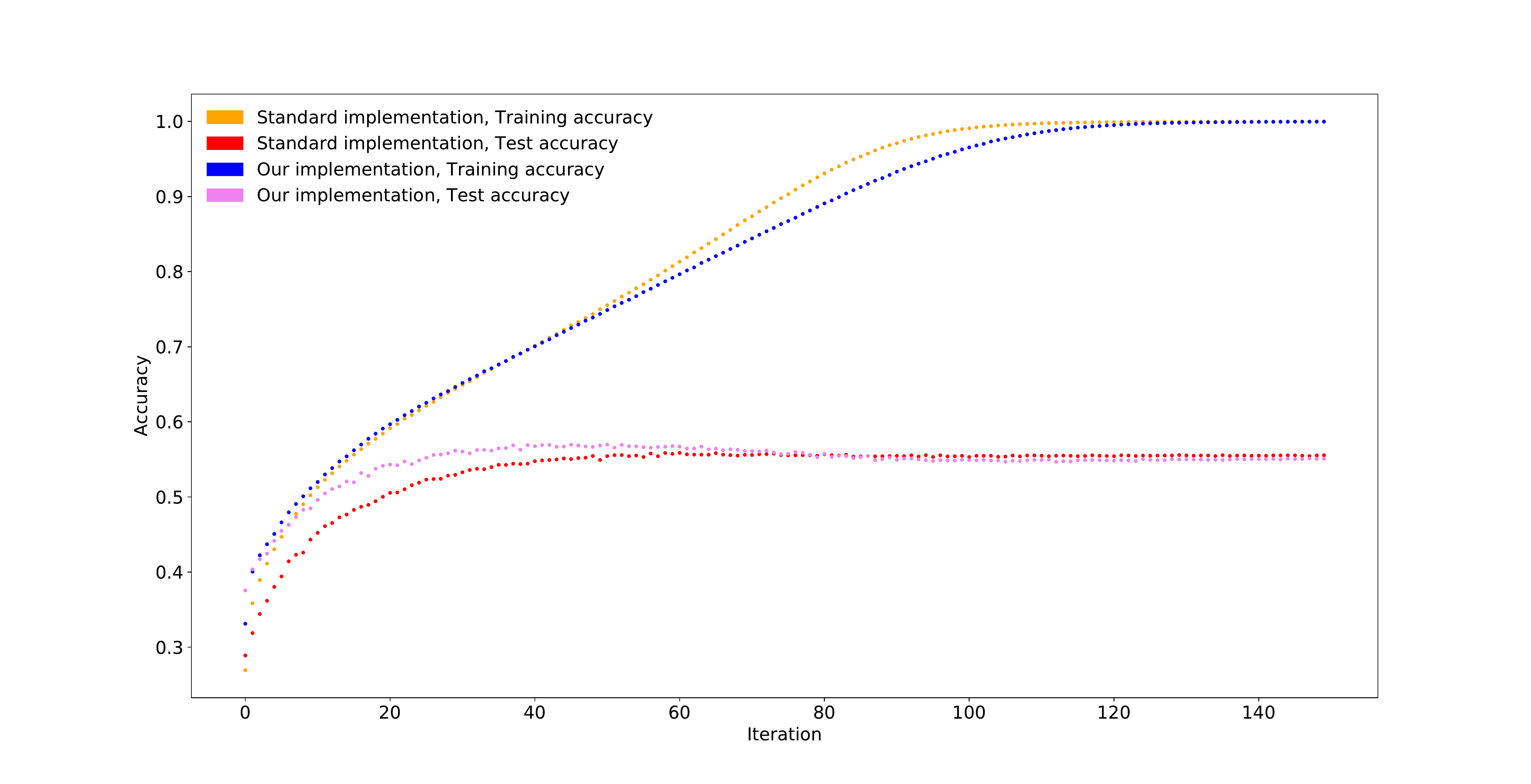}
\vspace*{-2mm}
\caption{Comparison of the average accuracy of a standard LeNet5 implementation with the one we generate, averaged over 10 training runs.}
\label{fig:FeatureNet-LeNet5-Accuracy}
\end{figure*}

To further check the validity of these architectures we semantically compare the LeNet5 architecture of our model (representation displayed in Figure \ref{fig:FeatureNet-LeNet5}) with the one implemented directly in Tensorflow/Keras (with the same parameters). Figure \ref{fig:FeatureNet-LeNet5-Accuracy} shows the training and test accuracy on CIFAR-10 (averaged over 10 runs). 
We observe that both types of accuracy follow the same trend in both models over the epochs. This confirms the conformance of our implementation. Here it must be noted that the slight recorded differences are inherent to the stochastic nature of the training process. 


\begin{figure*}
\centering
\includegraphics[width=0.9\linewidth]{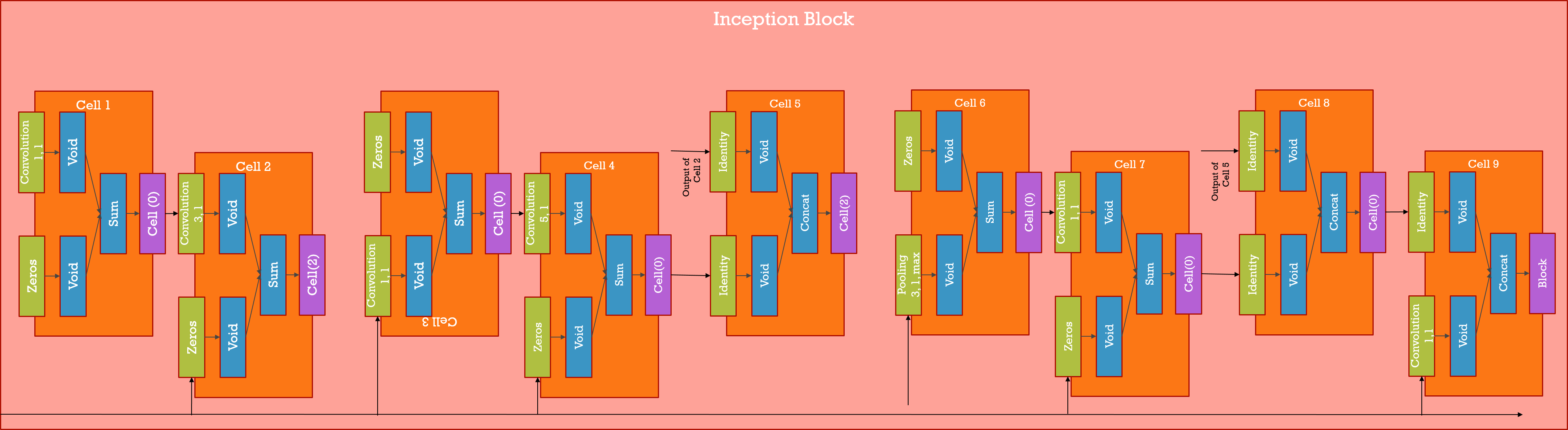}
\caption{An inception module (part of inception architecture) modelled using our representation.}
\label{fig:FeatureNet-InceptionModule}
\end{figure*}

Figure \ref{fig:FeatureNet-InceptionModule} shows an inception module modelled after our representation. This shows the capability of our modelling to support the branching and merging behaviour of such modules. Overall, an inception module concatenates 4 layers at once. We achieve this by performing 3 concatenations (in Cell 5, Cell 8 and Cell 9). For instance, Cell 5 concatenates the outputs from a convolution with kernel size 3 (Cell 2) and a convolution with kernel size 5 (Cell 4).








All in all, our results show that our FM can represent the space of DNN architectures and can successfully automate the generation of their variants. 

\subsection{RQ2: Searching the configuration space.}



\begin{figure*}
     \centering
     \begin{subfigure}[b]{0.45\linewidth}
         \centering
         \includegraphics[width=\linewidth]{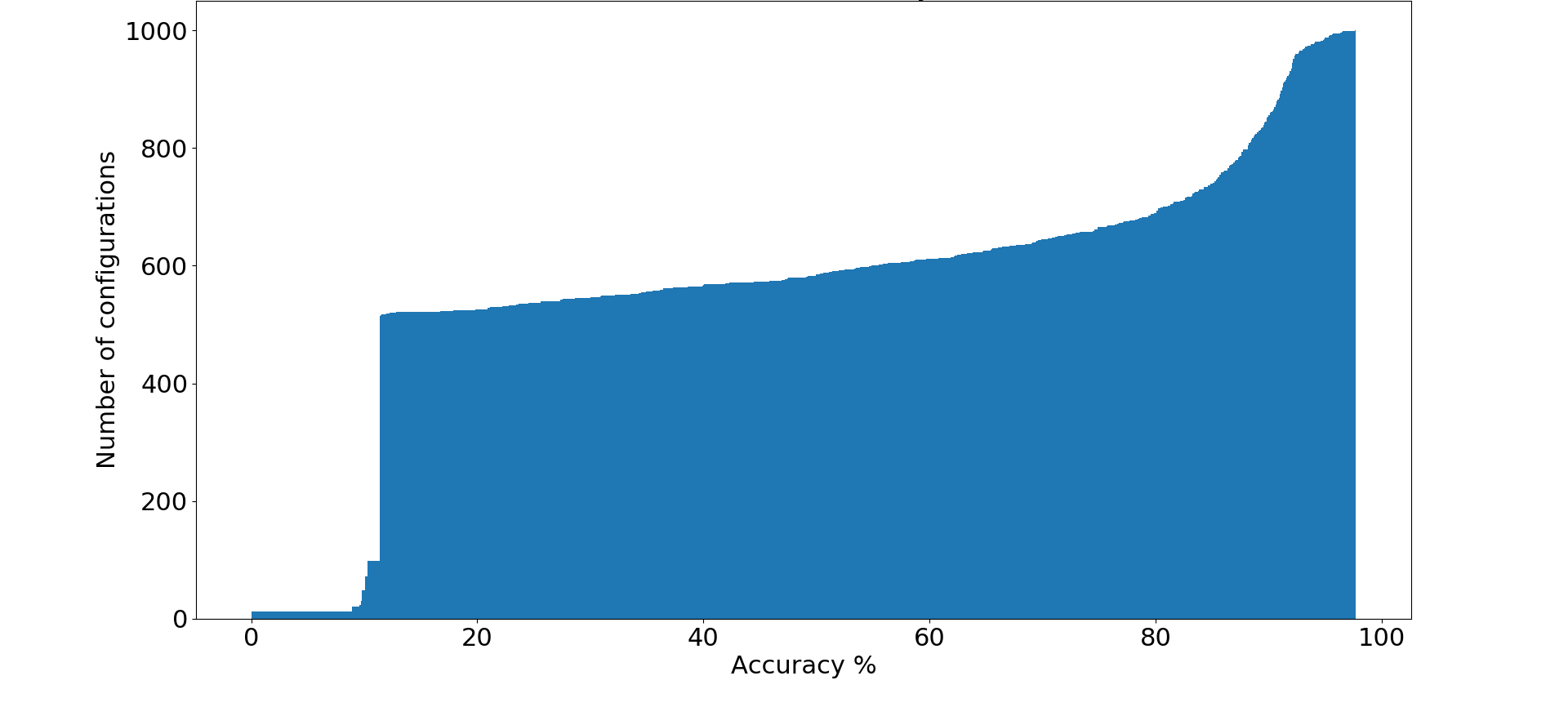}
         \caption{On MNIST}
         \label{fig:acc_1000_mnist}
     \end{subfigure}
     \begin{subfigure}[b]{0.45\linewidth}
         \centering
         \includegraphics[width=\linewidth]{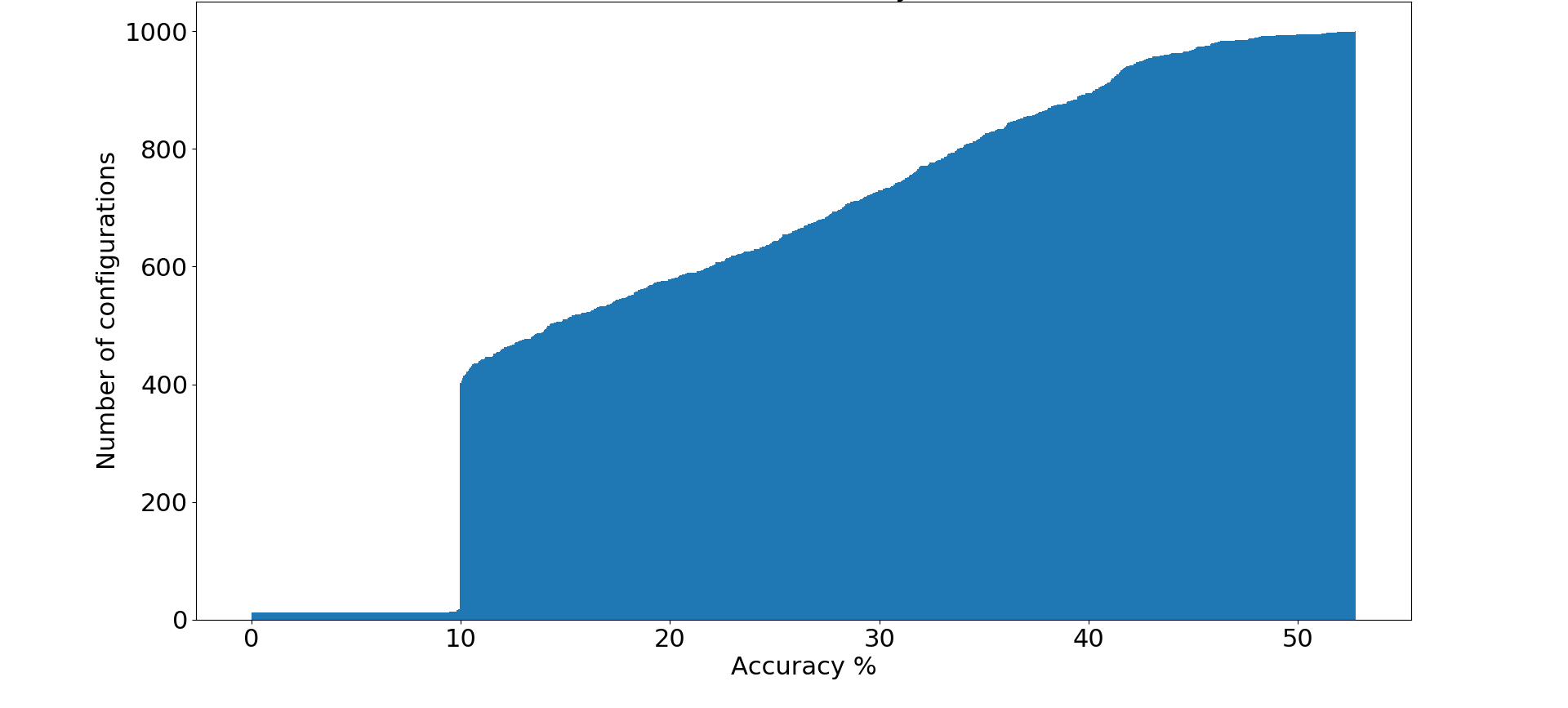}
         \caption{On Cifar-10}
         \label{fig:acc_1000_cifar}
     \end{subfigure}
    \caption{Distribution of the 1,000 generated architectures over all percentages of accuracy on two datasets. Any point $(x,y)$ of the graph denotes that $y$ architectures achieve an accuracy lower than $x\%$.}
        \label{fig:distribution_accuracy}
\end{figure*}

Figure \ref{fig:acc_1000_mnist} shows the accuracy values obtained by our configurations, when trained for 12 epochs. For each percentage of accuracy, we draw the number of architectures (out of the sample of 1,000) that achieve at most this accuracy percentage. For instance, we observe that 585/1000 architectures have an accuracy lower than 50\%, while that number jumps to 852 for 90\% accuracy. In the end, only 15 architectures (i.e. 1.5\% of the sample) achieves more than 95\% of accuracy. 

Figure \ref{fig:acc_1000_cifar} shows the results for CIFAR-10. We observe that our configurations are scattered over different ranges of accuracy indicating large performance differences. CIFAR-10 being more challenging than MNIST, 729 configuration barely reach 30\% of accuracy, while only 994 are below 50\%. Overall, none of the architectures manage to achieve 55\% of accuracy, due to our limitation to 12 epochs. Later in Section\ref{sec:results-rq3} we study how the best architectures obtained at 12 epochs perform when they are trained for more epochs.

\begin{figure*}
     \centering
     \begin{subfigure}[b]{0.45\linewidth}
         \centering
         \includegraphics[width=\linewidth]{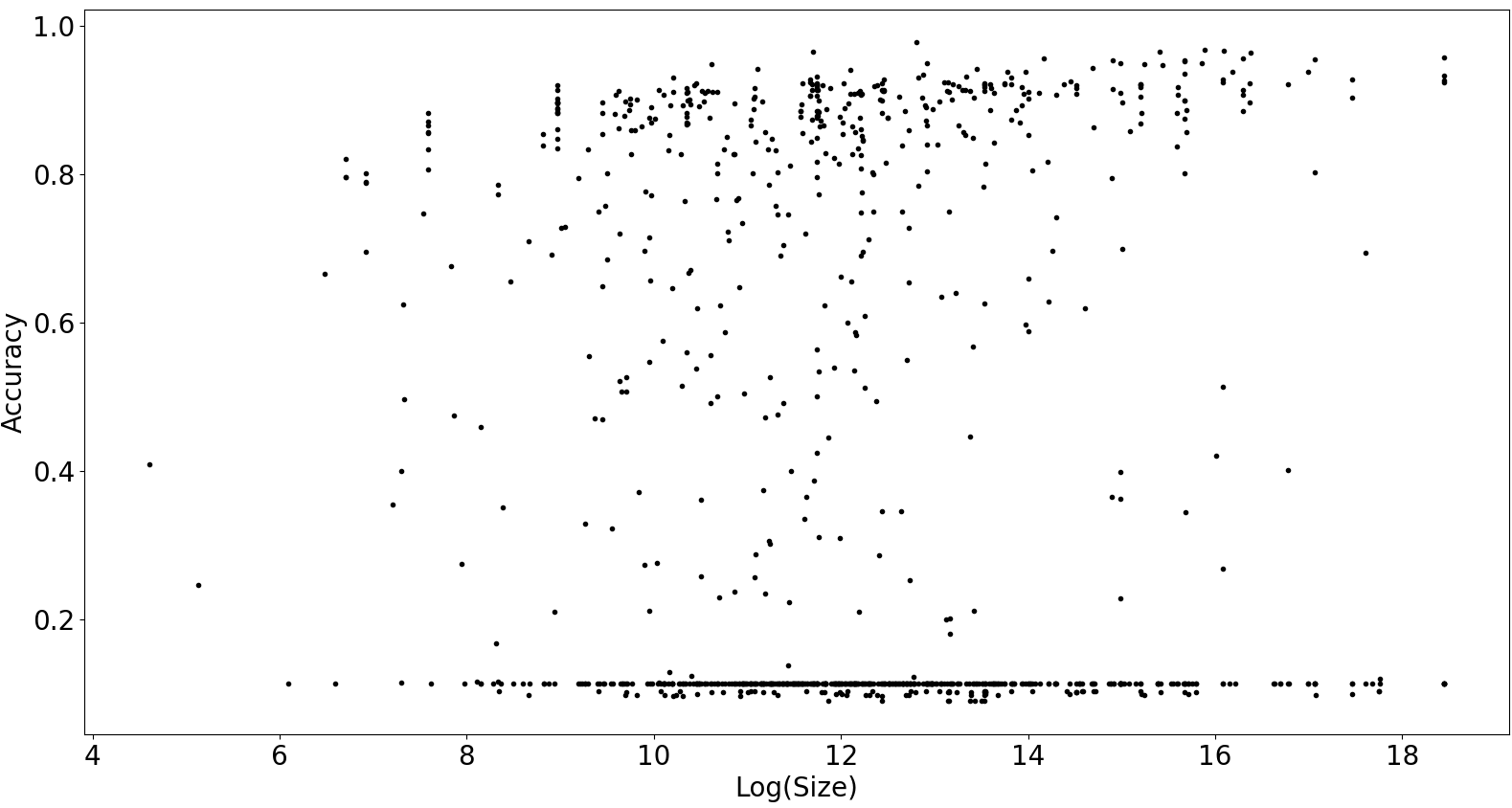}
         \caption{On MNIST}
         \label{fig:acc_1000_mnist_size}
     \end{subfigure}
     \begin{subfigure}[b]{0.45\linewidth}
         \centering
         \includegraphics[width=\linewidth]{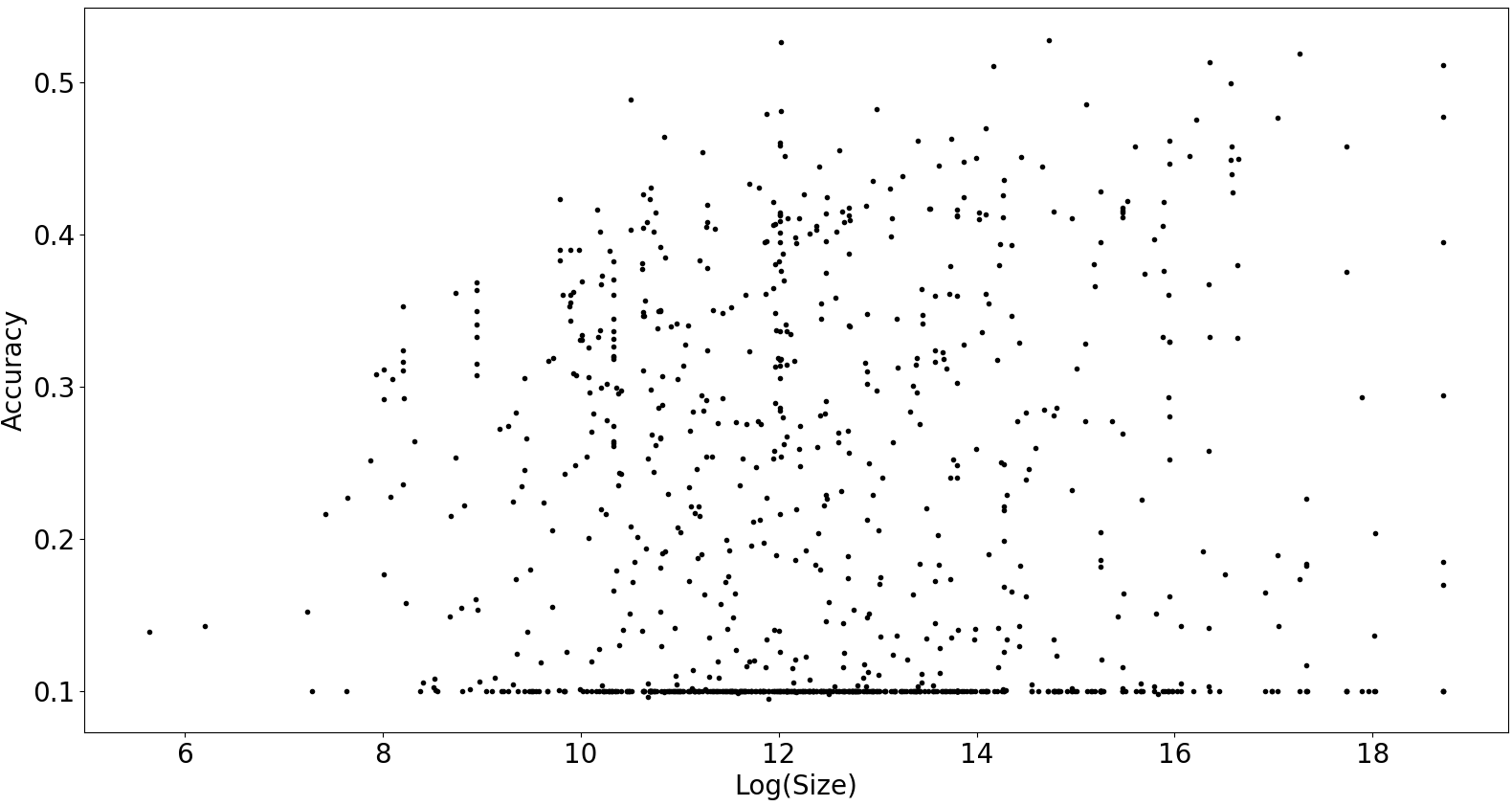}
         \caption{On Cifar-10}
         \label{fig:acc_1000_cifar_size}
     \end{subfigure}
    \caption{Distribution of the accuracy and size of 1,000 generated architectures. The size is given in terms of number of trainable weights of the architectures. By sampling a diverse set of configurations (in terms of feature differences), our technique is able to generate architectures with a wide range of sizes an few thousands to millions of weights. Moreover, we observe that architectures with high accuracy are not necessarily the ones with the largest size, while architectures performing poorly are found in all size ranges.}
        \label{fig:distribution_accuracy_size}
\end{figure*}

Overall, our results confirm that the DNN architectures have a substantial impact on accuracy, and that searching the configuration space leads to architectures that significantly outperform others.
A more detailed distribution of the accuracy and size of our 1000 configurations is detailed in Figure \ref{fig:distribution_accuracy_size}. We observe that architectures with high accuracy are not necessarily the ones with the largest size, while architectures performing poorly are found in all size ranges. This further supports the motivation for searching through the whole configuration space and not restricting it to large-sized architectures.

Taken together with the finding of RQ1, we can claim that our framework has the ability to form and specialize architectures for a particular domain, i.e., architectures that perform best in a particular domain. 



\subsection{RQ3: Finding architectures outperforming the state-of-the-art.}
\label{sec:results-rq3}

To conduct a fine-grained evaluation of our approach, we generate architectures three times under different constraints. The first sample, denoted S1, denotes the 1,000 architectures obtained by applying the general setup described previously, which thus cannot be substantially larger than LeNet5 as parameterized in \cite{LeCun1998GradientbasedLA}. The second one, denoted S2, restricts the sampling to 100 LeNet5 architectures with different parameterizations. Finally, the third one consists of 100 architectures and results from applying additional constraints to S2 that enforce smaller architectures: 5x5 convolution kernels are replaced by 3x3 kernels, convolution strides are increased from 1x1 to 3x3. In the following experiments, we retain the most accurate architectures obtained from the three samples, which we denote by \emph{Top S1}, \emph{Top S2} and \emph{Top S3}, respectively. Note that, in what follows, we use a slight abuse of language: by LeNet5, we refer to the LeNet5 architecture as it is parameterized in \cite{LeCun1998GradientbasedLA}. By contrast, S2 samples over the whole range of parameterizations of LeNet5 architectures.


\subsubsection{MNIST, 12 epochs}

The top part of Table \ref{table:3} shows the accuracy, size (in number of weights) and efficiency of LeNet5~\cite{LeCun1998GradientbasedLA} and our top sampled architectures, where $efficiency = \frac{accuracy \times 10^6}{size}$. We observe that, by selecting architectures of similar size to LeNet5, we obtain an architecture (\emph{Top S1}) which performs better than LeNet5. It is actually slightly more accurate than LeNet5 while being 34\% smaller, thereby leading to an increased efficiency. When we restrict the sampling to 100 LeNet5 architectures, the most accurate we obtained (\emph{Top S2}) is, overall, slightly more accurate and slightly bigger than LeNet5. \emph{Top S3} illustrates an interesting tradeoff between accuracy and size, as it is smaller than the other three by one order of magnitude, while still providing an accuracy of 92.31\%. This leads to an efficiency 12 times higher than LeNet5. We also observe that SqueezeNet performs poorly on MNIST. We see two reasons behind this. First, like most recent architectures, SqueezeNet is tailored to handle more challenging datasets (with larger and more detailed images) than MNIST and CIFAR-10. Second, it may require more training epochs to reach an acceptable accuracy.

\begin{table}
\centering
\caption{Accuracy, size and efficiency of LeNet5, SqueezeNet and our best sampled architectures, with a 12-epoch training, on MNIST (top part) and CIFAR-10 (bottom part).}
\begin{tabular}{r||r|r|r|r}
Dataset & Architecture & Accuracy & Size & Efficiency\\
\hline
\hline
\textbf{MNIST} & LeNet5 & 97.14\% & 545546 & 1.78 \\
& Top S1 &  97.74\% & 365194 &  2.68\\
& Top S2 & 97.65\% & 570218 & 1.71 \\
& Top S3 & 92.31\% & 43578 & 21.18 \\
& SqueezeNet & 43.67\% & 858154 & 0.51 \\
\hline
\textbf{CIFAR-10} & LeNet5 & 49.13\% & 868406 & 0.57 \\
& Top S1 & 52.77\% & 2494858 & 0.21 \\
& Top S2 & 57.79\% & 862646 & 0.67 \\
& Top S3 & 37.44\% & 38842 & 9.64 \\
& SqueezeNet & 17.96\% & 876970 & 0.51 \\
\end{tabular}
\label{table:3}
\end{table}


\subsubsection{CIFAR-10, 12 epochs}

Next, we repeat the same experiments on the CIFAR-10 dataset. We also consider the SqueezeNet architecture, which is known to offer an excellent efficiency on expensive datasets~\cite{Iandola2017SqueezeNetAA}. Results are shown in the bottom part of Table~\ref{table:3}. While \emph{Top S1} obtains a higher accuracy than LeNet5, this comes at the cost of a tripled size, thereby leading to a worst efficiency. However, by restricting the generation to LeNet5 architectures, we obtain an architecture (\emph{Top S2}) of similar size that offers a substantial increase in accuracy (from 49.13\% to 57.79\%). The architectures resulting from the third sample reach 37.44\% of accuracy at best, but exhibit a size more than 20 times smaller than LeNet5, which leads to a way higher efficiency. Like previously, SqueezeNet performs poorly on CIFAR-10. According to its inventors~\cite{Iandola2017SqueezeNetAA}, this architecture can achieve more than 90\% accuracy on CIFAR-10 by tweaking the training process (with momentum and decay), augmenting the data and applying various regularization methods. Such improvements are, however, out of the scope of our study as we are only focusing on the architecture structure and its parameters.

Overall, we conclude from both experiments that, with a 12-epoch training, our approach indeed manages to outperform an established architecture that was designed manually, be it on accuracy or efficiency, and that effective architectures are not necessarily the largest.




\subsubsection{CIFAR-10, 600 epochs}

To confirm that the above results are not due to the limited amount of epochs we allowed for training, we conducted additional experiments over more epochs. More precisely, we retain the 10 architectures resulting from the first generation S1 that have the highest accuracy. Then, we train them for a total of 600 epochs on the CIFAR-10 datasets, and compare the obtained performance metrics against LeNet5 and SqueezeNet. We limited the total training of each architecture to 100 minutes; only two of our architectures ended their training prematurely because of that threshold.

Table \ref{table:comparison_600epochs_cifar} shows the results at 12, 300 and 600 epochs, where the examined architectures are ordered by descending order of accuracy at 600 epochs. We observe that the gap between the best-performing architecture and LeNet5 tends to increase as more epochs are allowed for training (74.74\% against 59.26\%), which confirms the effectiveness of our approach to discover better architectures. 
Although the accuracy of SqueezeNet increases drastically over the epochs, it remains lower than most of the others, including LeNet5. Interestingly, the architecture  performing the best at 300 and 600 epochs was the worst one (if we except SqueezeNet) at 12 epochs. 

\begin{table}
\centering
\caption{Accuracy of the 10 best architectures from S1, LeNet5 and SqueezeNet on CIFAR-10 and at 12, 300 and 600 training epochs. Architectures are ordered by descending order of accuracy at 600 epochs. * indicates shortened training.}
\begin{tabular}{r|r|r|r|r}
Architecture  &  Size & (12) &  (300) & (600)\\
\hline
\#063 & 0.45M & 48.25\% & \textbf{74.28\%} & \textbf{74.74\%}  \\
\#203 & 0.17M & 52.64\% & 64.55\% & 65.25\% \\
\#161 & 3.62M & 48.57\% & 64.46\% & 65.24\% \\
\#477 & 2.49M & \textbf{52.77\%} & 63.43\% &  64.25\% \\
\#444 & 15.73M & 49.97\% & 62.40\% &  62.80\% \\
\#143 & 12.68M & 51.37\% & 60.46\% & 60.17\% \\
LeNet5 & 0.87M & 49.13\% & 59.42\% &  59.26\% \\
\#634 & 1.43M & 51.11\% & 59.76\% &  * 59.06\% \\
\#936 & 0.04M & 48.92\% & 54.26\% &  56.33\% \\
\#595 & 134.22M & 52.16\% & 52.64\% &  * 53.64\% \\
SqueezeNet & 0.88M & 17.96\% & 46.17\% &  49.69\% \\
\#059 & 31.51M & 51.9\% & 52.10\%  & 49.47\% \\
\end{tabular}

\label{table:comparison_600epochs_cifar}
\end{table}

\begin{figure*}
\centering
\hspace*{-0.8in}
\includegraphics[width=1.2\linewidth, trim={1cm 0.8cm 1cm 1cm},clip]{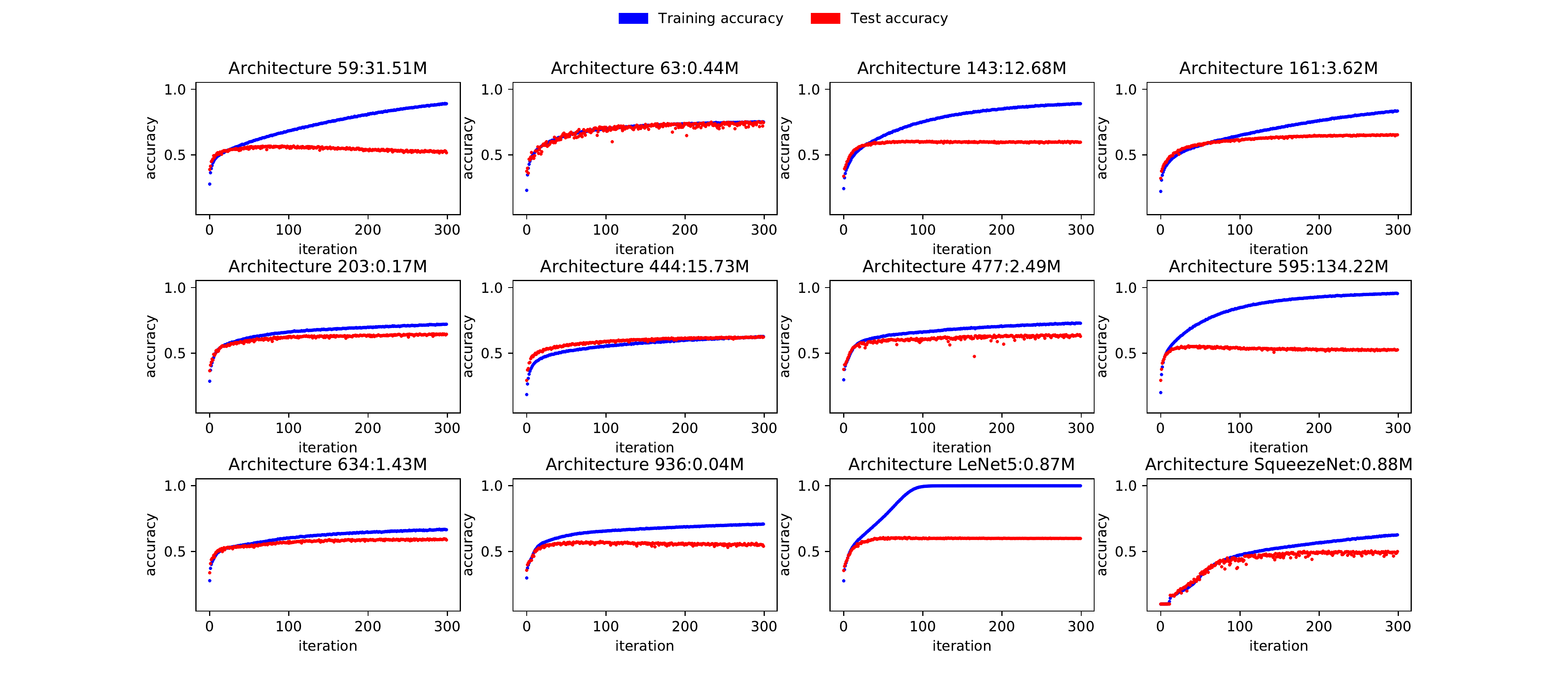}
\vspace*{-5mm}
\caption{Evolution of training and test accuracy over the training epochs. For each architecture, the title of the plot includes its size (in Million parameters) for reference}
\label{fig:acc_top_300}
\end{figure*}

We also see that, while the accuracy of all architectures increases from 12 to 300 epochs, four of the ten sampled ones have a slightly lower accuracy at 600 epochs than at 300 epochs, likely due to overfitting on the training set. To investigate this further, we show in Figure \ref{fig:acc_top_300} the evolution of the training accuracy (i.e. accuracy obtained by evaluating the architecture on the training set) and of the test accuracy (i.e. on the test set) for all the twelve architectures and from 1 to 300 epochs.


First, these results confirm that architecture size is not the main factor for accuracy and its evolution over the epochs. For example, \#143 and \#444 have comparable sizes but behave very differently, with \#143 overfitting more and more on the training data. Furthermore, the architectures where the gap between training and testing accuracy is the biggest are not necessarily the largest architectures.

Architectures \#063 and \#444 are the ones where training accuracy and test accuracy remain close and keep on increasing over the epochs. This is a good indicator that they can achieve even higher accuracy with more training. It also explains the fact that \#063 was relatively bad at 12 epochs while it becomes the best afterwards. It also means that some architectures from our initial 1,000 generated set that we discarded at 12 epochs could end up outperforming the final ten in the long run.

Conversely, other architectures, such as \#059 and \#936, see their test accuracy increase until approximately 50 epochs, before it decreases. If we look closely at LeNet5, we observe that it rapidly overfits on the training set as well. As a consequence, its test accuracy tends to stabilize at 50 epochs and does not seem to progress further. 

\subsection{Threats to Validity}


Validity threats may arise due our implementation. Our FM model might not be entirely accurate and our implementation might not be entirely correct. We do not consider this threat as major since we checked our models and implementations using a large sample of configurations. Moreover, our technique relies on widely-used frameworks, Tensorflow and Keras, and publicly available architectures (LeNet-5 and SqueezeNet) as a comparison baselines, all of which have stood to expert scrutiny. Overall, to reduce these threats we make our models and implementation publicly available.

Another threat regards the accuracy metrics we use on the two image datasets, MNIST and CIFAR-10. While accuracy is the commonly used metric, it does not reflect the extent to which our data architectures overfit. As we did not experiment with datasets involving a large variety of many or few classes and/or bigger images, it could be the case that our results are coincidental and do not generalize to other cases. Though, our framework manages to identify architectures leading to a large range of results indicating its ability to specialize to specific cases.  



Other threats may affect the generality of our framework. We exclusively focused on DNN architectures and eliminated all optimizations of the training process, such as data augmentation\cite{Simard2003BestPF}, modern regularization techniques (e.g. group-lasso regularizer \cite{GRPLassoRegularizer}, global average pooling, \cite{DBLP:journals/corr/LinCY13}, etc.), learning rate decays and improved stochastic gradient descent methods \cite{DBLP:journals/corr/Ruder16} (e.g. Adam \cite{Kingma2014}) in order to perform the experiment in a reasonable amount of time. All these innovations may impact both the accuracy and the generalization (to unseen data) of DNNs. By adding such mechanisms to our study, we can likely get better accuracy for both our generated models and the baseline ones. 


Another threat may arise due to our restrictions, i.e., we constraint the size of the architectures due to our limited computation resources. More powerful hardware and longer training times would be necessary to scout larger search spaces and more complex configurations. Still, our study already shows promising results on smaller search space and could be extended in the future to corroborate our conclusions on larger architectures. 


Finally, we compared our architectures with two references architectures, LeNet5 and SqueezeNet. Still, we consider them as representative because they form the current state-of-the-art. These architectures require a similar amount of weights to train on Cifar-10, about 0.88 millions, which is of the same scale as our best architecture (\#064). Besides, while LeNet5 is an old architecture that only uses the basic concepts of Convolution and Pooling, SqueezeNet which was proposed in 2017 uses a large variety of innovations, such as ReLU activations, Dropout, Skip connections, Convolution filters concatenation and a wide range of kernel sizes and number of features in each of its layers. SqueezeNet is therefore a good summary of the modern techniques proposed to improve image classification without massively increasing the size of the neural network. 

\section{Related Work}
\label{sec:related}

\subsection{Neural Architecture Search}

Designing a DNN is an iterative process that requires significant human expertise, research and computation time to tune the architecture and its hyperparameters to the target task and dataset, without any insurance that the resulting model will lead to acceptable results.


As of today, there is no consensus on a standard method to automatically generate these configurations. Automated Machine Learning (AutoML), the field that aims at providing engineers with automated means of building and deploying customized machine learning models, has therefore become a very important research topic. In particular, Neural Architecture Search (NAS) aims at identifying the best DNN architecture given specific tasks and performance metrics. Our work naturally falls into this area.

The time complexity of NAS approaches can be seen as $ O(N\bar{t})$ where $N$ is the number of assessed architectures and $\bar{t}$ the computation time required to assess every architecture. However, most techniques require a large $N$ to reach good performances. The main approach consists in hyperparameter optimization using reinforcement learning, with techniques like MetaQNN \cite{Baker2017DesigningNN}, ENAS \cite{pmlr-v80-pham18a} and DeepArchitect\cite{DeepArchitect}.  \\

MetaQNN uses Q-learning (a type of reinforcement learning) with Markov decision processes to iteratively select network layers and their parameters among a finite space. It defines every state as a layer with a set of hyperparameters, and is thus limited to the traditional view of DNNs, which we generalize in our representation to account for the latest architectures. More precisely, it supports only convolution, polling and dense layers, with a maximum of 2 dense layers and 12 overall. Another difference is the used search criterion: while theirs is based on the Q-learning metric, we rely on a sampling method driven by diversity. Although it achieves good performance, this method requires the evaluation of multiple thousands of architectures, leading to 8-10 days of calculation with 10 NVIDIA GPUs. 


Efficient Neural Architecture Search (ENAS), based on PNAS \cite{Liu2018ProgressiveNA}, is a meta-learning approach that uses a recurrent neural network to train and reinforce a target CNN architecture. It starts from a simple component (also named a cell) and stacks it $N$ times in order to build a final CNN. Their search problem is therefore reduced to identifying the best cell structure to replicate, rather than designing the entire convolutional network. 
This approach shares with our work the idea that we can achieve complex structures using a cellular approach. However, the components of their cells have no standard representation. An operator in ENAS cell can be applied on one or multiple inputs and can lead to multiple outputs. By contrast, our technique relies on a finer-grained configuration process where the standard structure of all cells and blocks is customizable independently.  


Finally, more flexible approaches to NAS are based on evolutionary algorithms (e.g. Genetic CNN \cite{Xie2017GeneticC}) but their performance and scalability are still to be assessed\cite{Zoph2018LearningTA}. 

\subsection{Configuration sampling}

In the context of Software Product Lines (SPLs), sampling is mainly used as an answer to the infeasibility of assessing every software product\cite{Varshosaz:2018:Samplingstudy}, due to the explosion of the configuration space as the number of features increases. While we considered diversity-based sampling  
\cite{Similarity2014}, other techniques are definitely of interest and should be evaluated in the near future.

Most immediate solutions rely on uniform/random sampling, which unfortunately cannot scale to large feature models~\cite{Plazar2019}. The extensive literature of product sampling includes many techniques to reduce the number of products to evaluate. Kim et al.\cite{Kim2011} propose a technique based on static software analysis, where irrelevant features (i.e. with no impact) are removed. Johansen et al.\cite{10.1007/978-3-642-33666-9_18} propose to split the SPL into sub-SPL to represent the knowledge of which interactions are prevalent.    

Sampling can also be driven by feature coverage. T-wise algorithms~\cite{Perrouin:2010:AST:1828417.1828490} aims to cover all $t$-uples of features (e.g. pair-wise aims to cover every pair) at least one. Such techniques become more expensive as $t$ increases, while they have to take into account the constraints of the feature model~\cite{Johansen:2012:AGT:2362536.2362547, Borazjany2012CombinatorialTO}.

Finally, multiple recent works combine techniques from machine learning and statistics to predict the performance of system configurations based on a sample (e.g. \cite{Guo2018}).


\section{Conclusion}
\label{sec:conclusion}

In this paper we demonstrated how to model DNN architectures with a FM. Such a modelling enables the application of variability management techniques on DNNs. Our variability model covers the most popular DNN architectures that have been engineered and hand-crafted in the past twenty years. Thanks to the compositional nature of FMs, our model can be extended easily to support more architectures. For instance, recurrent neural networks\cite{Sherstinsky2018FundamentalsOR} have introduced a looping mechanism to allow reasoning over sequences of inputs, which is notably useful for speech recognition and language modelling. Modelling such non-linear networks would require allowing cells' output to reach a preceding cell or block.

The hierarchical structure of FMs also allows the addition of new dimensions of variability. While the current FM focuses on the inner constituents of DNN architectures, we could extend it to include variability in the training setup (e.g. hyperparameters like learning rate, optimization method, data augmentation techniques and etc.) or in user preferences (performance metrics, explainability and etc.), and allow the possibility to restrict the configuration to a specific family of architectures (e.g., LeNet) through the dynamic addition of constraints.

An additional contribution of our paper regards the fully automated process that searches and deploys DNN architectures. Our search procedure relies on an out-of-the-box diversity-based configuration generation, which leads to architectures that outperform the state-of-the-art. Of course, our framework is not limited to diversity-based generation, which is used for demonstration purposes, and can be extended with other configuration generation techniques. 
Nevertheless, since the training of hundreds of architectures demands tremendous computation resources, future research should focus on experimenting with additional search techniques, such as SATIBEA~\cite{Henard2015}, or developing specialized techniques potentially combined with simulation and performance prediction methods \cite{abs-1801-02175} that have the potential to significantly reduce the required computations.

The framework, the FM and datasets used in this paper are available at:
\textit{\url{https://github.com/yamizi/FeatureNet}}

\bibliographystyle{ACM-Reference-Format}
\bibliography{splc19,mcr}

\end{document}